\documentclass{article}
\usepackage{spconf,amsmath,graphicx}

\usepackage{subfig}
\usepackage{bm}
\usepackage{float}
\usepackage{amsfonts}
\usepackage{diagbox}
\usepackage{color}
\usepackage{cite}
\def\M{\mathrm{M}}
\def\FF{\mathrm{FF}}
\def\Q{\mathrm{Q}}
\def\K{\mathrm{K}}
\def\V{\mathrm{V}}
\def\H{\mathrm{H}}

\title{ON THE USEFULNESS OF SELF-ATTENTION FOR AUTOMATIC SPEECH RECOGNITION WITH TRANSFORMERS}
\name{Shucong Zhang, Erfan Loweimi, Peter Bell, Steve Renals}
\address{Centre for Speech Technology Research, University of Edinburgh, Edinburgh, UK}
\begin{document}
%\ninept
%
\maketitle
\begin{abstract}
Self-attention models such as Transformers, which can capture temporal relationships without being limited by the distance between events, have given competitive speech recognition results. However, we note the range of the learned context increases from the lower to upper self-attention layers, whilst acoustic events often happen within short time spans in a left-to-right order. This leads to a question: for speech recognition, is a global view of the entire sequence useful for the upper self-attention encoder layers in Transformers?  To investigate this, we train models with lower self-attention/upper feed-forward layers encoders on Wall Street Journal and Switchboard. Compared to baseline Transformers, no performance drop but minor gains are observed. We further developed a novel metric of the diagonality of attention matrices and found the learned diagonality indeed increases from the lower to upper encoder self-attention layers.  We conclude the global view is unnecessary in training upper encoder layers.  
\end{abstract}
\section{Introduction}

Self-attention networks (SANs) have recently become a popular research topic in the speech recognition community \cite{Dong2018Speech-transformer, zhou2018syllable, pham2019veryDeep, povey2018time, zeyer2019comparison, wang2019transformerHybrid, karita2019comparative, nakatani2019improving,lu2020exploring} and they can yield superior results compared to recurrent neural networks (RNNs), which are conventionally used to model sequential data. However, due to gradient vanishing, it is difficult for RNNs to model long-range dependencies \cite{bengio1994learning}, even with gated structures such as Long Short-Term Memory (LSTM)~\cite{hochreiter1997LSTM} and Gated Recurrent Unit (GRU)~\cite{chung2014GRU}. In SANs, self-attention layers encode contextual information through attention mechanisms \cite{bahdanau2015neural, vaswani2017attention}. With this mechanism, when learning the hidden representation for each time step of a sequence, a self-attention layer has a global view of the entire sequence and  thus can capture temporal relationships without the limitation of range. This is believed to be a key factor for the success of SANs \cite{vaswani2017attention}. 

%In this paper we study Transformers~\cite{vaswani2017attention}, end-to-end SAN-based models with two components: an encoder and a decoder. The encoder uses self-attention layers to encode input sequences. At decoding time step $t$, the decoder generates the current output by attending to the encoded input sequence and to the outputs generated before time $t$. For attention-based RNN end-to-end models \cite{luong2015effective, bahdanau2015neural}, the RNN encoder encodes the input sequence. The RNN decoder interacts with the encoded input sequence through an attention layer to produce outputs. 

Previous works on attention-based RNN end-to-end models has shown that for speech recognition, since acoustic events usually happen in a left-to-right order within small time spans, restricting the attention to be monotonic along the time axis improves the model's performance \cite{tjandra2017local, kim2017joint, zhang2019windowed}. This appears to be in contrast to the reason for the success of SANs: if the global view provided by the attention module of self-attention layers is beneficial,  why then does forcing the attention mechanism to focus on local information result in performance gains for RNN end-to-end models?

To investigate this, we study Transformers~\cite{vaswani2017attention}, which are end-to-end SAN-based models. We explore training Transformers with upper (further from the input) feed-forward layers and lower self-attention layers encoders. The feed-forward layers can be viewed as  ``monotonic left-to-right diagonal attention''. We performed extensive experiments on the Wall Street Journal  (WSJ) read speech corpus  \cite{paul1992wsj} and  the Switchboard (SWBD) conversational telephone  speech corpus \cite{godfrey1992switchboard}, finding that the upper feed-forward layers do not lead to higher error rates -- they even give improved accuracy. 

To further analyse each self-attention layer, we have developed a novel metric for the diagonality of attention matrices. Based on this metric we found that the overall trend of the average diagonality of each layer increases from the lower layers to the upper layers. Thus, even given a global view of inputs, the upper layers learned to only attend local information during training. The lower layers, on the other hand, have learned to captured long range context through the self-attention mechanism.

These observations resolves the seemingly contradiction between the previous studies on RNN-based end-to-end models which restrict the attention to be diagonal and the reason for the success of SAN-based models.  For attention-based RNN models, the attention mechanism interacts with both the decoder and the encoder. Since an output unit (e.g. a character) is often related to a short time span of acoustic features, the attention layer should attend to a small window 
%of the elements in 
of the encoded input sequence in a left-to-right order. In this work we study a self-attention encoder which learns the hidden representation for each time step of the input sequence. The global view of the input sequences enables the lower layers to encode context information well. When the lower layers capture sufficient contextual information, is the self-attention mechanism not useful for the upper layers. Thus, we conclude the upper self-attention layers are not useful and they can be replaced by feed-forward layers. 

 \section{Related Work}
Self-attention and its multi-head attention module \cite{vaswani2017attention} which uses multiple attention mechanisms to encode context are key components of Transformers. Michel et al~\cite{michel2019are} remove a proportion of the heads in the multi-head attention for each self-attention layer in trained Transformers, finding it leads to minor performance drops. This implies that not all the attention heads are equally useful. In our work, instead of removing some attention heads in trained models, we replace entire self-attention layers with a feed-forward layers and train models with feed-forward layers as the upper layers in the encoder .

For a self-attention layer, a single-layer feed-forward module is stacked on the multi-head attention module. Irie et al~\cite{irie2020how} extend the single-layer feed-forward module to a multi-layer module, arguing it can bring more representation power, and show that a SAN with fewer modified self-attention layers (as well as fewer parameters) can have minor performance drops compared to a SAN with a larger number of the original self-attention layers. In this work we study the effect of the stacked context among the self-attention layers of the encoder.  We do not change the architecture of the self-attention layers and we replace the upper self-attention layers in the encoder of Transformers with feed-forward layers.

Previous works have investigated restricting each self-attention layer to attend a small window of context and observed a decrease in accuracy \cite{wang2019transformerHybrid,lu2020exploring}. In this work we observed that the lower self-attention layers tend to learn a larger window of context compared to the upper layers. Thus assigning a uniform window length to each layer may not be optimal. The upper feed-forward lower self-attention layers encoders can be viewed as imposing a window of length one for the upper layers, without restricting the window length for the lower layers.

When the upper self-attention layers are replaced with feed-forward layers, the architecture of the encoder is similar to the CLDNN (Convolutional, Long Short-Term Memory Deep Neural Network) \cite{sainath2015CLDNN}. The CLDNN uses an LSTM to model the sequential information and a deep neural network (DNN) to learn further abstract representation for each time step. Stacking a DNN on an LSTM results in a notable error rate reduction compared to pure LSTM models. While we found the upper self-attention layers of the encoder of Transformers can be replaced with feed-forward layers, stacking more feed-forward layers does not result in further performance gains. The main goal of this work is to understand the self-attention encoder.

\begin{figure}[!htb]
\centering
\subfloat[SA]{\includegraphics[width=0.50\linewidth]{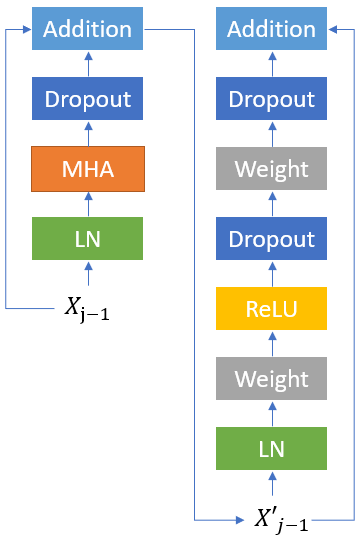}} \quad
\subfloat[FF]{\includegraphics[width=0.222\linewidth]{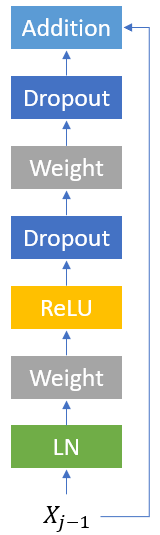}}
\caption{Architectures of a self-attention (SA) encoder layer with multi-head attention (MHA) and a feed-forward (FF) encoder layer. LN is layer normalization \cite{ba2016layer}. We omit LN and dropout \cite{srivastava2014dropout} in the equations of encoder layers but they are applied in the experiments.  } 
\label{fig:FF}
\end{figure}

\section{Model Architecture}
%In this section we describe multi-head attention, self-attention layers, and the self-attention encoder module in Transformers. Then, we introduce the replacement of the upper self-attention layers of the encoder by feed-forward layers. 

\subsection{Multi-head Attention}
Multi-head attention uses attention mechanisms to encode sequences \cite{vaswani2017attention}. We firstly consider a single attention head. The input sequences to the attention mechanism are mapped to a query sequence $\bm{Q}$, a key sequence $\bm{K}$, a value sequence $\bm{V}$ where $\bm{K}$ and $\bm{V}$ have the same length. For the $i$-th element $\bm{Q}[i]$ of $\bm{Q}$, an attention vector is generated by computing the similarity between $\bm{Q}[i]$ and each element of $\bm{K}$. Using the attention vector as weights, the output is a weighted sum over the value sequence $\bm{V}$. Thus, an attention head $\mathrm{A}$ of the multi-head attention can be described as:
\begin{align}
    \mathrm{A}(\bm{X}^\Q, \bm{X}^\K, \bm{X}^\V) &= \mathrm{softmax}(\frac{\bm{Q} \bm{K}^T }{\sqrt{d^{\mathrm{K}}}})\bm{V}\\
(\bm{Q} ,\bm{K} ,\bm{V}) &= (\bm{X}^\Q \bm{W}^\Q, \bm{X}^{\K}\bm{W}^{\K},\bm{X}^\V \bm{W}^\V),
\end{align}
where $\bm{X}^\Q \in \mathbb{R}^{ n \times d^{\M}}$, $\bm{X}^\K, \bm{X}^\V \in \mathbb{R}^{ m \times d^{\M}}$ are inputs and $m,n$ denote the lengths of the input sequences; $\bm{W}^\Q, \bm{W}^\K \in \mathbb{R}^{ d^{\M} \times d^{\K}} $ and  $\bm{W}^\V \in \mathbb{R}^{ d^{\M} \times d^{\V}}$ are trainable matrices. The three input sequences $(\bm{X}^\Q, \bm{X}^\K, \bm{X}^{\V})$ can be the same sequence, e.g., the speech signal to be recognised. The multi-head attention $\mathrm{MHA}$ uses $h$ attention heads $(A_{1},A_{2},\cdots,A_{h})$ and a trainable matrix $\bm{U}^\H \in \mathbb{R}^{d^{\H} \times d^{\M}}, d^{\H} =h \times d^{\V}$ to combined the outputs of each attention head:
\begin{equation}
    \mathrm{MHA}(\bm{X}^\Q, \bm{X}^\K, \bm{X}^\V) = (A_{1},A_{2},\cdots,A_{h}) \ \bm{U}^\H
\end{equation}

\subsection{Self-attention Encoder}
The self-attention encoder in a Transformer is a stack of self-attention layers. The $j$-th 
% encoder self-attention 
layer reads the output sequence $\bm{X}_{j-1}$ from its lower layer and uses multi-head attention to process the input sequence. That is,  $(\bm{X}^Q, \bm{X}^K, \bm{X}^V)= (\bm{X}_{j-1},\bm{X}_{j-1},\bm{X}_{j-1})$. The multi-head attention only contains linear operations. Thus, in a self-attention layer, a non-linear feed-forward layer is stacked on the multi-head attention module. A self-attention layer in the encoder of a Transformer can be described as:
\begin{align}
    \bm{X}'_{j-1} &= \bm{X}_{j-1} + \mathrm{MHA}(\bm{X}_{j-1},\bm{X}_{j-1},\bm{X}_{j-1}) \\
    \bm{X}_{j} &= \bm{X}'_{j-1} + \mathrm{ReLU}(\bm{X}'_{j-1}\bm{S} +\bm{b})\bm{Z}+\bm{r}
\end{align}
where $\bm{S} \in \mathbb{R}^{d^{\M}\times d^{\FF}}$, $\bm{Z} \in \mathbb{R}^{d^{\FF} \times d^{\M}}$, $\bm{b} \in \mathbb{R}^{d^{\FF}}$ and $\bm{r} \in \mathbb{R}^{d^{\M}}$ are trainable matrices and vectors . 

%\vfil
%\qquad

\subsection{Feed-Forward Upper Encoder Layers}
 In the encoder, since each self-attention layer learns contextual information from its lower layer, the span of the learned context increases from the lower layers to the upper layers. Since acoustic events often happen within small time spans in a left-to-right order, if the inputs to the upper layer have encoded a sufficient large span of context, then it is unnecessary for the upper layers to learn further temporal relationships. Thus, the multi-head attention module which extracts the contextual information could be redundant, and the self-attention layer will not be essential.  
However, if  the upper layers of the encoder are self-attention layers and the lower layers have already seen a sufficiently wide context, then the attention mechanism will focus on a narrow range of inputs, since no further contextual information is required. Assuming that acoustic events often happen left-to-right, the attention matrix will tend to be diagonal. Then, since $\mathrm{MHA}(\bm{X}_{j-1},\bm{X}_{j-1},\bm{X}_{j-1})\approx \bm{X}_{j-1}$ and self-attention is not helpful,  replacing self-attention layers with feed-forward layers will not lead to a drop in accuracy. 

The architecture of the feed-forward layers is:
\begin{equation}
    \bm{X}_{j} = \bm{X}_{j-1} + \mathrm{ReLU}(\bm{X}_{j-1}\bm{S} +\bm{b}) \ \bm{Z}+\bm{r}
\end{equation}
Figure~\ref{fig:FF} demonstrates the architecture of a self-attention layer and a feed-forward layer. Furthermore, a feed-forward layer can be viewed as a self-attention layer with an identity matrix as its attention matrix.

%Note that if the self-attention is not useful for the upper layers, then replacing them with feed-forward layers will not lead to accuracy drops; however, observing feed-forward layers yield no drop in accuracy  does not imply that upper self-attention is redundant in the base model. It is possible that for the base model, the contextual information is still useful for the upper layers. When the upper layers are replaced by the feed-forward layers, these feed-forward layers cannot learn contextual information and thus they force their lower layers to learn a larger window of context as compensation. However if the essential context is captured by the lower layers, the upper layers do not need further context information and feed-forward networks are sufficient for the upper layers. We distinguish these two situations in Section~\ref{sec: diag}. 
%Compared to the self-attention layer, a feed-forward layer also has fewer parameters and reduces decoding time. 

%\vspace{-4mm}
\section{Experiments and Discussion}
%\vspace{-2mm}
\subsection{Experimental Setup}
%\vspace{-2mm}

We experiment on two datasets, Wall Street Journal (WSJ) which contains 81 hours of read speech training data and Switchboard (SWBD), which contains 260  hours of  conversational telephone speech training data. We use WSJ dev93 and eval92  test sets and SWBD  eval2000 and SWBD/callhome test sets. We use Kaldi \cite{povey2011kaldi} for data preparation and feature extraction -- 83-dim log-mel filterbank frames with pitch \cite{ghahremani2014pitch}. The output units for the WSJ experiments are 26 characters, and the  apostrophe, period, dash, space, noise and \emph{sos} / \emph{eos} tokens. The output tokens for SWBD experiments are tokenized using Byte Pair Encoding (BPE)~\cite{sennrich2016BPE}.

We compare Transformers with different types of encoders. The baseline Transformer encoders comprise self-attention layers and are compared with Transformers whose encoders have feed-forward layers following the self-attention layers. Each self-attention/feed-forward layer is counted as a single layer, and encoders with the same number of layers are compared. All the components of each model have the same architecture, except for the number of self-attention/feed-forward layers in the encoder,

We employ 12-layer encoders, since a 12-layer architecture is consistent with previous works and has been widely used for Transformer models \cite{Dong2018Speech-transformer, karita2019comparative,michel2019are, lu2020exploring,nakatani2019improving}. We also test 6-layer encoders for the WSJ dataset. Other settings of the models follow  \cite{karita2019comparative}.

In each model, below the Transformer's encoder there are two convolutional neural network layers with 256 channels, with a stride of 2 and a kernel size of 3, which map the dimension of the input sequence to $d^{\M}$. The multi-head attention components of the self-attention layers have 4 attention heads and $d^{\V}=d^{\K}=64$, $d^{\M}=256$. For the feed-forward module of the self-attention layers, as well as for the proposed feed-forward encoder layers, $d^{\FF}=2048$. Dropout rate $0.1$ is used when dropout is applied. The Transformer decoder has 6 layers. Input sequences to the encoder and the decoder are concatenated with sinusoidal positional encoding \cite{vaswani2017attention}. Models are implemented using  ESPnet  \cite{watanabe2018espnet} and PyTorch \cite{paszke2017automatic}.

The training schedule (warm up steps/learning rate decay) follows \cite{Dong2018Speech-transformer}. Adam \cite{kingma2015adam} is used as the optimizer. The batch size is 32. Label smoothing with smoothing weight 0.1 is used. We train the model for 100 epochs and the averaged parameters of the last 10 epochs are used as the parameters of the final model \cite{Dong2018Speech-transformer}. Besides the loss from the Transformer's decoder $L^{\mathrm{D}}$, a connectionist temporal classification (CTC) \cite{graves2006connectionist} loss $L^{\mathrm{CTC}}$ is also applied to the Transformer encoder \cite{kim2017joint}. Following the previous work \cite{karita2019comparative}, the final loss $L$ for the model is: 
\vspace{-2mm}

\begin{equation}
    L = (1-\lambda)L^{\mathrm{D}} + \lambda L^{\mathrm{CTC}}
\end{equation}
where $\lambda = 0.3$ for WSJ and $\lambda = 0.2$ for SWBD.
\begin{figure}[tb]
\centering
\subfloat[Attention vectors of each attention head of encoder self-attention layer 12 ]{\includegraphics[width=1.0\columnwidth]{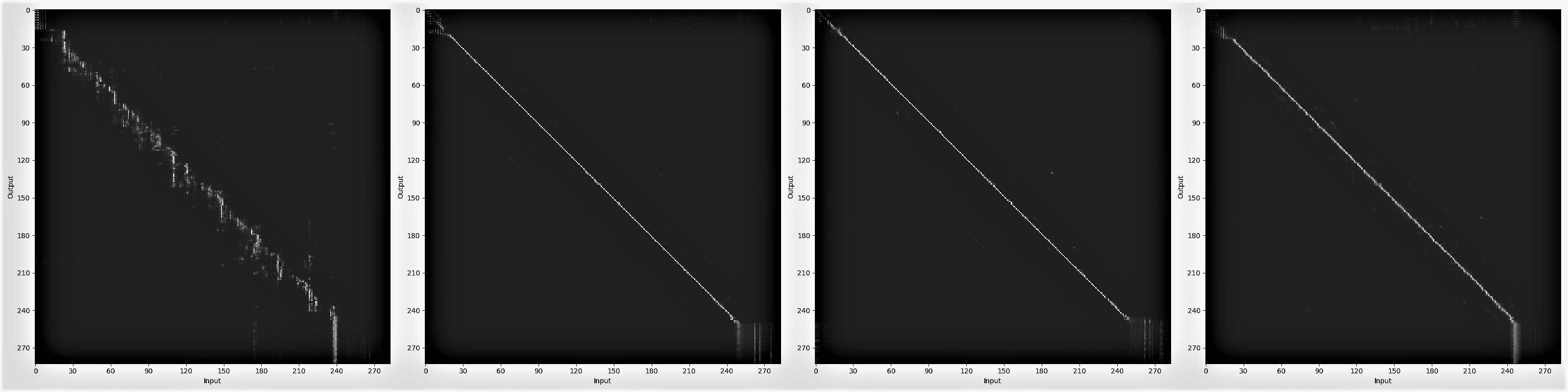}}
\quad
\subfloat[Attention vectors of each attention head of encoder self-attention layer 5 ]{\includegraphics[width=1.0\columnwidth]{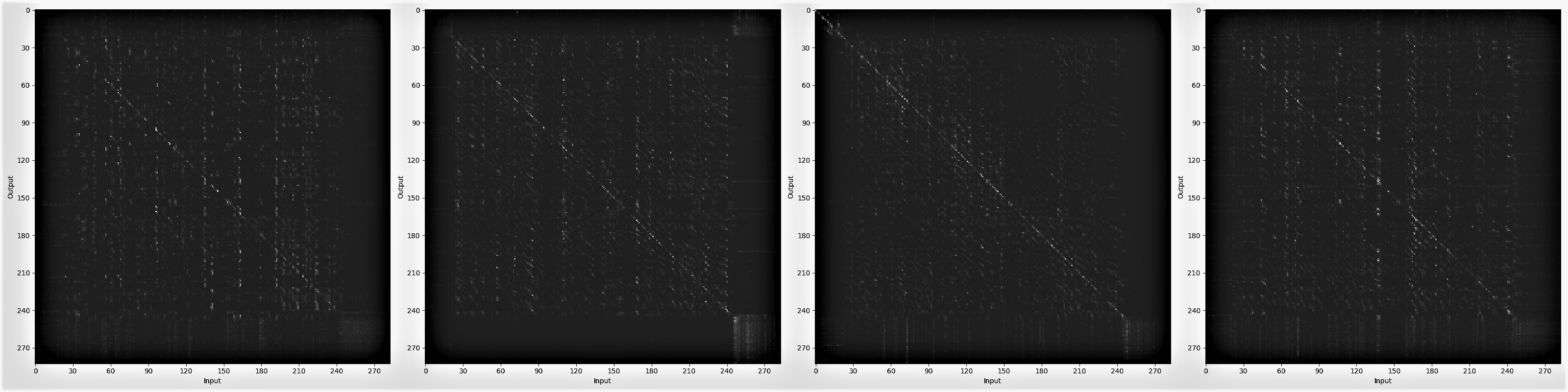}}
\quad
\subfloat[Attention vectors of each attention head of encoder self-attention layer 1]{\includegraphics[width=1.0\columnwidth]{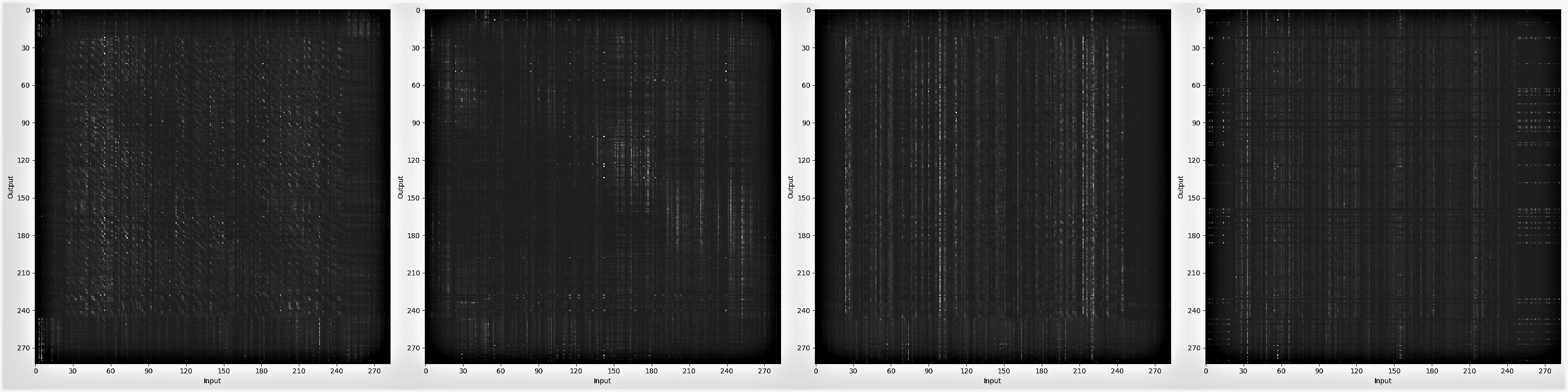}}

\caption{A sample of attention vectors of encoder self-attention layers generated by the baseline Transformer with a 12-layer encoder. The sampled utterance is form WSJ eval92. While the lowest layer (layer1, near input) attends a wide range of context, the middle layer focus more on the local information and the topmost layer assigns nearly all the attention weight to the diagonal.} 
\label{fig:attention}
\end{figure}

\subsection{Experimental Results on WSJ}

For the experiments on WSJ, we first train a baseline model with a 12-layer self-attention encoder. Then, we use this model to decode WSJ eval92 and  compute the attention matrices of a randomly sampled utterance from eval92.  Figure~\ref{fig:attention} shows the plots of the  attention matrices for each attention head of the lowest layer, a middle layer and the highest layer. The lowest layer attends to a wider range of context. The middle layers put more attention weight on the diagonal and the middle two heads of the topmost layer have close to pure diagonal attention matrices which can be described as $\mathrm{MHA}(\bm{X}_{j-1},\bm{X}_{j-1},\bm{X}_{j-1})\approx \bm{X}_{j-1}$. This implies even given a global view of inputs during training, the topmost layer learned to only focus on local information. Section~\ref{sec: diag} discusses the statistics of the ``diagonality'' of  attention matrices for each head of every layer.

\begin{table}[ht]
\caption{Character error rate (CER) on WSJ for the Transformer models with different encoders. The evaluation sets are WSJ eval92 and dev93. SA denotes self-attention layer and FF denotes feed-forward layer.}
\label{tab:wsj1}
\centering
\begin{tabular}{l|l|l|l|l}
\hline
\multicolumn{3}{l|}{Number of Layers}               & \multicolumn{2}{l}{CER/\%}    \\ \hline \hline
Total & SA & FF & eval92      & dev93     \\ \hline \hline
12 & 12             & 0            & 3.5          & 4.6          \\
12 & 11             & 1            & \textbf{3.4} & \textbf{4.5} \\
12 & 10            & 2            & 3.6          & 4.6          \\
12 & 9              & 3            & 3.8          & 4.8          \\
12 & 8              & 4            & 3.9          & 4.9          \\
12 & 7              & 5            & 4.0          & 5.1          \\
12 & 6              & 6            & 4.2          & 5.3          \\ \hline \hline
11 & 11            & 0            & 3.6          & 4.7          \\
10 &  10           & 0            & 4.0          & 5.2          \\ \hline \hline
13 & 12             & 1            & 3.6          & 4.7          \\
13 & 11             & 2            & 3.6          & 4.6          \\
14 & 11            & 3            & 3.7          & 4.6          \\ \hline \hline
6 &       6       & 0            & 4.2          & 5.4          \\
6  & 5           & 1            & 4.2          & \textbf{5.3}          \\
6 & 4              & 2            & \textbf{4.1}          & 5.6          \\
6 & 3             & 3            & 4.4          & 5.9          \\
\hline
\end{tabular}
\end{table}

After training the baseline, we train models whose encoders are built by different numbers of self-attention layers and feed-forward layers. For the encoder of these models, there are 12 layers in total and the lower layers are self-attention layers while the upper layers are feed-forward layers. We start from an encoder with 6 self-attention layers and 6 feed-forward layers. Then, we increase the number of self-attention layers and decrease the number of feed-forward layers. Table~\ref{tab:wsj1} shows that as the number of self-attention layers increases, the character error rate (CER) decreases, which implies learning further contextual information is beneficial.

However, when the number of self-attention layers increases to 10, with 2 upper feed forward layers, the encoder gives almost identical results compared to the 12-layer self-attention baseline, although the 10-layer self-attention encoder has notably higher CERs. Furthermore, although the 11-layer self-attention encoder gives worse results compared to the 12-layer baseline, the encoder which has 11 self-attention layers and one upper feed forward layers yields the best results. Increasing the number of self-attention layers to 12 and decreasing the number of feed forward layers to 0 is harmful. This set of experiments shows it is crucial for the layers below the $10^{th}$ layer to encode temporal relationships. Upon the $10^{th}$ layer the global view of the sequence is not useful, indicating the contextual information is well captured by the layers beneath.

We further tested if stacking more feed-forward layers to make deeper encoders is beneficial. As shown in Table~\ref{tab:wsj1}, this does not give performance gains. We also investigated modifications to the architecture of the stacked feed-forward layers, such as removing  residual connections or using an identity mapping \cite{he2016identity}. These modifications  did not result in a  CER reduction compared to the 11-layer self-attention 1-layer feed-forward encoder.

We also tested the 6-layer encoder architecture and the results are also shown in Table~\ref{tab:wsj1}. The baseline model has 6 self-attention layers as its encoder. Then we replace the top one, two and three layers with feed-forward layers respectively. We observe that replacing the topmost layer of the 6-layer self-attention encoder does not lead to reductions in accuracy but to minor improvements, which is consistent with the experimental results for the 12-layer encoder. 

%Indeed, this observation is also interesting. For the 12-layer encoder, the 6$^{th}$ layer is a middle layer. Replacing the self-attention layers above the 6$^{th}$ leads to performance drops, which implies contextual information is still important for these layers. However, for the 6-layer encoder, the 6$^{th}$ layer does not behave like the  6$^{th}$ layer of the 12-layer encoder. Rather, as the topmost layer of the 6-layer encoder, it only encoders a small span of contextual information and can be replaced with feed-forward layers. 

\subsection{Experimental Results on SWBD}

We further test replacing upper self-attention layers on the  larger and more challenging SWBD corpus. The results are shown in Table~\ref{tab:SWBD}. The encoder with 10 self-attention layers is less accurate than the encoders with 11 and 12 self-attention layers. Also, the 12-layer self-attention encoder has higher word error rates (WERs) than the 11-layer encoder. However, the encoder with 10 self-attention layers and 2 feed-forward layers, which has 12 layers in total, gives the lowest WERs. The 9 self-attention layers + 3 feed-forward layers encoder yields higher WERs. Thus, the layers below the 10$^{th}$ layer is crucial in learning contextual information. Upon the 10$^{th}$ self-attention layer feed forward layers are sufficient in learning further abstract representations.

\begin{table}[tb]
\caption{Word error rate (WER) of the experiments on SWBD for the Transformer models with different encoders. The evaluation sets are eval 2000 SWBD/callhome. SA denotes self-attention layer and FF denotes feed forward layer.}
\label{tab:SWBD}
\centering
\begin{tabular}{l|l|l|l|l}
\hline
\multicolumn{3}{l|}{Number of Layers} & \multicolumn{2}{l}{WER/\%} \\ \hline \hline
Total & SA & FF & SWBD      & Callhome     \\ \hline \hline
12    & 12             & 0            & 9.0       & 18.1         \\ \hline
12    & 11             & 1            & 9.0       & 17.8         \\ \hline
12    & 10             & 2            & \textbf{8.9}       & \textbf{17.6}         \\ \hline
12    & 9              & 3            & 9.5       & 18.5         \\ \hline \hline
11    & 11             & 0            & 9.0       & 17.7         \\ \hline
10    & 10             & 0            & 9.2       & 18.4         \\ \hline \hline
\multicolumn{3}{l|}{Transformer\cite{karita2019comparative}}      & 9.0       & 18.1         \\ \hline
\multicolumn{3}{l|}{Transformer\cite{pham2019veryDeep}}      & 10.4      & 18.6         \\ \hline
\multicolumn{3}{l|}{Transformer\cite{zeyer2019comparison}}      & 10.6      & 22.3         \\ \hline
\end{tabular}
\end{table}

\subsection{Metric of Diagonality for Attention Matrices}
To further analyse each attention layer and each attention head, we propose a novel metric for the diagonality of attention matrices. The $j^{th}$ element in the $i^{th}$ row of the attention matrix is the attention weight between the $i^{th}$ element and the $j^{th}$ element of the input sequence of the self-attention layer. The attention weights sum to 1 in each row and the attention vector can be viewed as a probability distribution over each row. In the $i^{th}$ row, if all the probability mass is allocated to the $i^{th}$ element then it indicates, that for this row, all the attention weight is on the diagonal of the attention matrix. When the probability mass is assigned to be as far as possible from the $i^{th}$ element in the $i^{th}$ row for all rows, then the attention matrix has the lowest diagonality. Based on this, we first define the \emph{centrality} $C_i$ of row $i$:
\begin{equation}
C_i = 1-\frac{\sum_{j=1}^{n} a_{ij}\left | i-j \right |}{\mathrm{Max} (\left | i-1 \right |,\left | i-2 \right |,\cdots,\left | i-n \right |)  }
\end{equation}
where $j$ denotes the index of each column, $n$ denotes the length of the input sequence, $a_{ij}$ denotes the attention weight between the $i^{th}$ element and the $j^{th}$ element of the input sequence, and $\left | i-j \right |$ is the distance between the $i^{th}$ element and the $j^{th}$ element of the input sequence. Based on this definition, consider the first row of a $5 \times 5$ attention matrix. For such a matrix, $(1,0,0,0,0)$ will have centrality 1, $(0,0,0,0,1)$ will have centrality 0, and $(0.2,0.2,0.2,0.2,0.2)$ will have centrality 0.5. We define the \emph{diagonality} $D$ of an attention matrix as the average over the centrality of all its rows:
\begin{equation}
    D = \frac{\sum_{i=1}^{n} C_i}{n}.
\end{equation}

% \usepackage{multirow}

\iffalse
\begin{table}[!tb]
\caption{The mean diagonality with $\pm$ standard deviation for all attention heads of each layer for the 12-layer encoder. The topmost row denotes the topmost layer. }
\label{tab:12 diagonality}
\begin{tabular}{l|l|l|l}
\hline
head 1               & head 2              & head 3               & head 4               \\ \hline \hline
.906 $\pm$ .041 & .951$\pm$ .028 & .917 $\pm$ .044 & .926 $\pm$ .026 \\ \hline
.850 $\pm$ .037 & .885$\pm$ .032 & .710 $\pm$ .043 & .883 $\pm$ .043 \\ \hline
.938 $\pm$ .028 & .814$\pm$ .060 & .772 $\pm$ .041 & .678 $\pm$ .055 \\ \hline
.815 $\pm$ .038 & .838$\pm$ .034 & .657 $\pm$ .035 & .911 $\pm$ .036 \\ \hline
.930 $\pm$ .025 & .769$\pm$ .027 & .721 $\pm$ .030 & .953 $\pm$ .021 \\ \hline
.966 $\pm$ .022 & .970$\pm$ .016 & .705 $\pm$ .027 & .733 $\pm$ .026 \\ \hline
.686 $\pm$ .026 & .631$\pm$ .027 & .982 $\pm$ .009 & .987 $\pm$ .009 \\ \hline
.662 $\pm$ .031 & .695$\pm$ .031 & .740 $\pm$ .021 & .625 $\pm$ .033 \\ \hline
.697 $\pm$ .030 & .645$\pm$ .024 & .634 $\pm$ .032 & .712 $\pm$ .027 \\ \hline
.608 $\pm$ .029 & .613$\pm$ .032 & .636 $\pm$ .035 & .606 $\pm$ .032 \\ \hline
.627 $\pm$ .033 & .661$\pm$ .056 & .618 $\pm$ .035 & .841 $\pm$ .031 \\ \hline
.533 $\pm$ .038 & .568$\pm$ .034 & .612 $\pm$ .035 & .588 $\pm$ .040 \\ \hline
\end{tabular}
\end{table}
\fi

\begin{figure}[!tb]
\centering
\subfloat[12-layer Encoder]{\includegraphics[width=0.455\linewidth]{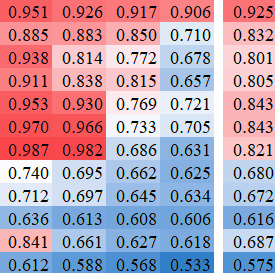}} \qquad
\subfloat[6-layer Encoder]{\includegraphics[width=0.455\linewidth]{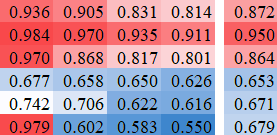}}
\caption{The heat map of the averaged diagonality of each attention head in each layer. The $5^{th}$ column shows the average diagonality over all heads of each layer. The red color denotes high diagonality and the blue color indicates low diagonality. }
\label{fig:heat map}
\end{figure}

\begin{figure}[!tb]
\centering
\includegraphics[width=1.0\columnwidth]{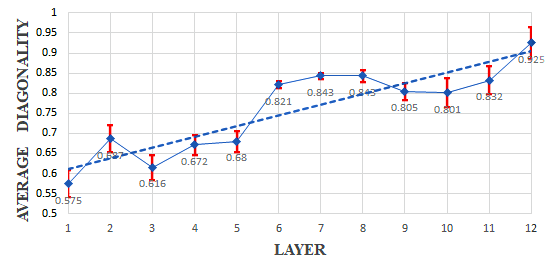}
\caption{The averaged diagonality with $\pm$ standard deviation of each self-attention layer of the 12-layer encoder baseline. Layer 12 is the topmost layer. The dash line is the trend line.}
\label{fig:d_12}
\end{figure}

\subsection{Diagonality of Each Layer}
\label{sec: diag}
To further evaluate the usefulness of self-attention for each layer, we compute the average diagonality of each attention head for every layer of the baseline 12-layer encoder model on WSJ eval92, and the average diagonality over all attention heads of each layer. As shown in Figure~\ref{fig:d_12}, the overall trend of the average diagonality indeed increases from the lower layers to the upper layers. In the experiments on replacing self-attention layers,  models with more than 2 feed-forward layers and fewer than 10 self-attention layers yield higher error rates (Table~\ref{tab:wsj1}). Figure~\ref{fig:d_12} shows average diagonality from the $9^{th}$ layer to the $10^{th}$ layer is relatively low, compared to the topmost two layers. These consistent observations indicate contextual information is necessary for the $9^{th}$ layer and the $10^{th}$ layer and thus the self-attention mechanism is essential for these two layers. For the topmost two layers, even with the self-attention mechanism, the diagonality is close to 1, which shows they focus on local information. This is also consistent with the finding in Table~\ref{tab:wsj1} that replacing these self-attention layers with feed-forward layers leads to no increase in error rate.

\begin{figure}[!tb]
\centering
\includegraphics[width=1\columnwidth]{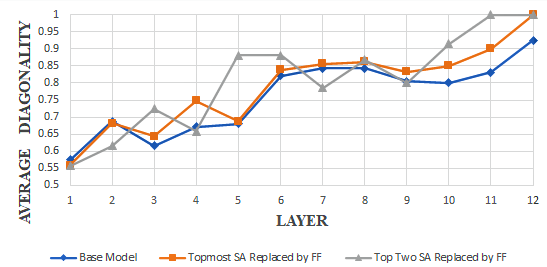}
\caption{The averaged diagonality of each layer of the encoders. Layer 12 is the topmost layer. Feed-forward layers have diagonality 1.}
\label{fig:d_12_FF}
\end{figure}

Another interesting observation is the average diagonality of the the $7^{th}$  and $8^{th}$ layers is also high. Thus, it is possible that self-attention is also not useful for these two layers. The reason for the high CERs of replacing the $7^{th}$ to the $12^{th}$ self-attention layers with feed-forward layers (Table~\ref{tab:wsj1}) could be the global view of the $9^{th}$  and  $10^{th}$ layers. We propose that  layers could be replaced not only based on their position but also based on their diagonality, such as replacing the $7^{th}$, $8^{th}$, $11^{th}$ and $12^{th}$ layer with feed-forward layers and leaving the $9^{th}$ and $10 ^{th}$ layer with self-attention.

The average diagonality of each attention head of every layer is shown in Figure~\ref{fig:heat map}. From the $6^{th}$ layer to the $8^{th}$ layer the diagonality of each head varies significantly -- two heads have diagonality close to 1 and two heads have relatively low diagonality.  These heads with high diagonality  are candidates for replacement with diagonal attention (feed-forward networks). To investigate how the diagonality changes after the upper layers are replaced by feed forward layers, we compute the average diagonality of each layer of the models with one and two feed-forward layers in the encoder, where the performance does not drop. Figure~\ref{fig:d_12_FF} shows the overall trend of the average diagonality still increases from the lower layers to the upper layers.

\iffalse
\begin{table}[!tb]
\caption{The mean diagonality with $\pm$ standard devation for all attention heads of each layer of the 6-layer encoder. The topmost row denotes the topmost layer. }
\label{tab:6 diagonality}
\begin{tabular}{l|l|l|l}
\hline
head 1               & head 2              & head 3               & head 4               \\ \hline \hline
.905 $\pm$ .028 & .814$\pm$ .023 & .831 $\pm$ .049 & .936 $\pm$ .025 \\ \hline
.984 $\pm$ .006 & .911$\pm$ .023 & .935 $\pm$ .020 & .970 $\pm$ .011 \\ \hline
.817 $\pm$ .028 & .970$\pm$ .009 & .801 $\pm$ .022 & .868 $\pm$ .023 \\ \hline
.658 $\pm$ .021 & .650$\pm$ .015 & .677 $\pm$ .018 & .625 $\pm$ .022 \\ \hline
.622 $\pm$ .023 & .616$\pm$ .014 & .706 $\pm$ .019 & .742 $\pm$ .004 \\ \hline
.550 $\pm$ .044 & .979$\pm$ .005 & .583 $\pm$ .013 & .602 $\pm$ .017 \\ \hline
\end{tabular}
\end{table}
\fi
\begin{figure}[!tb]
\centering
\includegraphics[width=1\columnwidth]{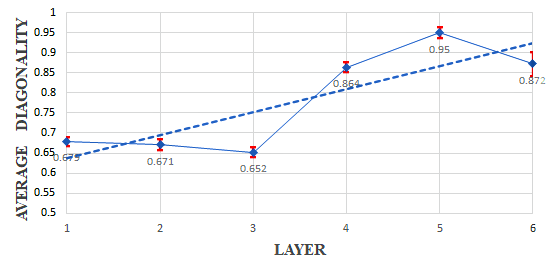}
\caption{The averaged diagonality with $\pm$ standard deviation of each self-attention layer of the 6-layer encoder baseline. Layer 6 is the topmost layer. The dash line is the trend line.}
\label{fig:d_6}
\end{figure}

\begin{figure}[!tb]
\centering
\includegraphics[width=1\columnwidth]{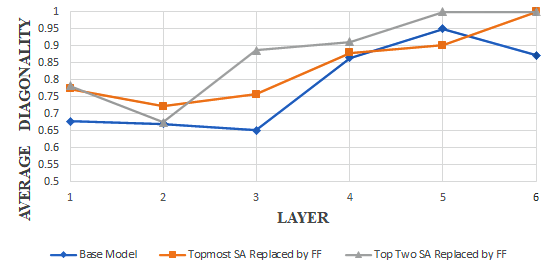}
\caption{The averaged diagonality of each layer of the encoders. Layer 6 is the topmost layer. Feed-forward layers have diagonality 1.}
\label{fig:d_6_FF}
\end{figure}

We also computed the average diagonality of each layer of the baseline 6-layer Transformer encoder. Figure~\ref{fig:d_6} shows the trend of the average diagonality increases from the lower layers to the upper layers. After replacing the top one or two layers of the self-attention layers with feed-forward layers, the trend of the average diagonality remains increasing (Figure~\ref{fig:d_6_FF}). For the baseline 6-layer encoder, the $5^{th}$ layer achieves the highest diagonality. Thus it is potential that only replacing the $5^{th}$ layer will not harm the performance of the model. Also, as Figure~\ref{fig:heat map} shows, the first layer of the 6-layer encoder has a head with a diagonality of $0.979$, which is clearly an outlier among the heads in the first layer, and a candidate for replacement with a feed-forward network. 

\section{Conclusion}
In this paper, based on the argument that acoustic events often happen in short time spans with a  left-to-right ordering, and that the encoded context increases through the lowest self-attention layer to the highest self-attention layer through the Transformer encoder, we investigate the usefulness of self-attention for the upper layers in the encoder. Our experiments on WSJ and SWBD show that replacing the upper self-attention layers with feed-forward layers does not increase the model's error rate. We developed a novel metric for the diagonality of the attention matrix, finding the overall diagonality indeed increases from the lower layers to the upper layers. These observations imply the self-attention is not useful for the upper layers of the encoder. Further work includes replacing self-attention heads and self-attention layers based on their diagonality and designing novel network architecture based on our findings.

\bibliographystyle{IEEEbib}
\bibliography{strings,refs}

\begin{thebibliography}{10}

\bibitem{Dong2018Speech-transformer}
Linhao Dong, Shuang Xu, and Bo~Xu,
\newblock ``Speech-transformer: a no-recurrence sequence-to-sequence model for
  speech recognition,''
\newblock in {\em 2018 IEEE International Conference on Acoustics, Speech and
  Signal Processing (ICASSP)}. IEEE, 2018, pp. 5884--5888.

\bibitem{zhou2018syllable}
Shiyu Zhou, Linhao Dong, Shuang Xu, and Bo~Xu,
\newblock ``Syllable-based sequence-to-sequence speech recognition with the
  transformer in mandarin chinese,''
\newblock {\em Proc. INTERSPEECH 2018}, pp. 791--795, 2018.

\bibitem{pham2019veryDeep}
Ngoc-Quan Pham, Thai-Son Nguyen, Jan Niehues, Markus M{\"u}ller, and Alex
  Waibel,
\newblock ``Very deep self-attention networks for end-to-end speech
  recognition,''
\newblock {\em Proc. INTERSPEECH 2019}, pp. 66--70, 2019.

\bibitem{povey2018time}
Daniel Povey, Hossein Hadian, Pegah Ghahremani, Ke~Li, and Sanjeev Khudanpur,
\newblock ``A time-restricted self-attention layer for asr,''
\newblock in {\em 2018 IEEE International Conference on Acoustics, Speech and
  Signal Processing (ICASSP)}. IEEE, 2018, pp. 5874--5878.

\bibitem{zeyer2019comparison}
Albert Zeyer, Parnia Bahar, Kazuki Irie, Ralf Schl{\"u}ter, and Hermann Ney,
\newblock ``A comparison of {Transformer} and {LSTM} encoder decoder models for
  {ASR},''
\newblock in {\em IEEE ASRU}, 2019.

\bibitem{wang2019transformerHybrid}
Yongqiang Wang, Abdelrahman Mohamed, Duc Le, Chunxi Liu, Alex Xiao, Jay
  Mahadeokar, Hongzhao Huang, Andros Tjandra, Xiaohui Zhang, Frank Zhang,
  et~al.,
\newblock ``Transformer-based acoustic modeling for hybrid speech
  recognition,''
\newblock {\em arXiv preprint arXiv:1910.09799}, 2019.

\bibitem{karita2019comparative}
Shigeki Karita, Nanxin Chen, Tomoki Hayashi, Takaaki Hori, Hirofumi Inaguma,
  Ziyan Jiang, Masao Someki, Nelson Enrique~Yalta Soplin, Ryuichi Yamamoto,
  Xiaofei Wang, et~al.,
\newblock ``A comparative study on transformer vs rnn in speech applications,''
\newblock {\em arXiv preprint arXiv:1909.06317}, 2019.

\bibitem{nakatani2019improving}
Tomohiro Nakatani,
\newblock ``Improving transformer-based end-to-end speech recognition with
  connectionist temporal classification and language model integration,''
\newblock {\em Proc. INTERSPEECH 2019}, 2019.

\bibitem{lu2020exploring}
Liang Lu, Changliang Liu, Jinyu Li, and Yifan Gong,
\newblock ``Exploring transformers for large-scale speech recognition,''
\newblock {\em arXiv preprint arXiv:2005.09684}, 2020.

\bibitem{bengio1994learning}
Yoshua Bengio, Patrice Simard, Paolo Frasconi, et~al.,
\newblock ``Learning long-term dependencies with gradient descent is
  difficult,''
\newblock {\em IEEE transactions on neural networks}, vol. 5, no. 2, pp.
  157--166, 1994.

\bibitem{hochreiter1997LSTM}
Sepp Hochreiter and J{\"u}rgen Schmidhuber,
\newblock ``Long short-term memory,''
\newblock {\em Neural computation}, vol. 9, no. 8, pp. 1735--1780, 1997.

\bibitem{chung2014GRU}
Junyoung Chung, Caglar Gulcehre, KyungHyun Cho, and Yoshua Bengio,
\newblock ``Empirical evaluation of gated recurrent neural networks on sequence
  modeling,''
\newblock {\em arXiv preprint arXiv:1412.3555}, 2014.

\bibitem{bahdanau2015neural}
Dzmitry {Bahdanau}, Kyunghyun {Cho}, and Yoshua {Bengio},
\newblock ``Neural machine translation by jointly learning to align and
  translate,''
\newblock in {\em ICLR 2015 : International Conference on Learning
  Representations 2015}, 2015.

\bibitem{vaswani2017attention}
Ashish Vaswani, Noam Shazeer, Niki Parmar, Jakob Uszkoreit, Llion Jones,
  Aidan~N Gomez, {\L}ukasz Kaiser, and Illia Polosukhin,
\newblock ``Attention is all you need,''
\newblock in {\em Advances in neural information processing systems}, 2017, pp.
  5998--6008.

\bibitem{tjandra2017local}
Andros Tjandra, Sakriani Sakti, and Satoshi Nakamura,
\newblock ``Local monotonic attention mechanism for end-to-end speech and
  language processing,''
\newblock in {\em Proceedings of the Eighth International Joint Conference on
  Natural Language Processing (Volume 1: Long Papers)}, 2017, pp. 431--440.

\bibitem{kim2017joint}
Suyoun Kim, Takaaki Hori, and Shinji Watanabe,
\newblock ``Joint ctc-attention based end-to-end speech recognition using
  multi-task learning,''
\newblock in {\em 2017 IEEE International Conference on Acoustics, Speech and
  Signal Processing (ICASSP)}. IEEE, 2017, pp. 4835--4839.

\bibitem{zhang2019windowed}
Shucong Zhang, Erfan Loweimi, Peter Bell, and Steve Renals,
\newblock ``Windowed attention mechanisms for speech recognition,''
\newblock in {\em 2019 IEEE International Conference on Acoustics, Speech and
  Signal Processing (ICASSP)}. IEEE, 2019, pp. 7100–--7104.

\bibitem{paul1992wsj}
Douglas~B Paul and Janet~M Baker,
\newblock ``The design for the wall street journal-based csr corpus,''
\newblock in {\em Proceedings of the workshop on Speech and Natural Language}.
  Association for Computational Linguistics, 1992, pp. 357--362.

\bibitem{godfrey1992switchboard}
J.J. {Godfrey}, E.C. {Holliman}, and J.~{McDaniel},
\newblock ``Switchboard: telephone speech corpus for research and
  development,''
\newblock in {\em 1992 IEEE International Conference on Acoustics, Speech and
  Signal Processing (ICASSP)}, 1992, pp. 517--520.

\bibitem{michel2019are}
Paul {Michel}, Omer {Levy}, and Graham {Neubig},
\newblock ``Are sixteen heads really better than one,''
\newblock in {\em NeurIPS 2019 : Thirty-third Conference on Neural Information
  Processing Systems}, 2019, pp. 14014--14024.

\bibitem{irie2020how}
Kazuki {Irie}, Alexander {Gerstenberger}, Ralf {Schluter}, and Hermann {Ney},
\newblock ``How much self-attention do we need? trading attention for
  feed-forward layers,''
\newblock in {\em 2020 IEEE International Conference on Acoustics, Speech and
  Signal Processing (ICASSP)}, 2020.

\bibitem{sainath2015CLDNN}
Tara~N. {Sainath}, Oriol {Vinyals}, Andrew {Senior}, and Hasim {Sak},
\newblock ``Convolutional, long short-term memory, fully connected deep neural
  networks,''
\newblock in {\em 2015 IEEE International Conference on Acoustics, Speech and
  Signal Processing (ICASSP)}, 2015, pp. 4580--4584.

\bibitem{ba2016layer}
Jimmy~Lei Ba, Jamie~Ryan Kiros, and Geoffrey~E Hinton,
\newblock ``Layer normalization,''
\newblock {\em arXiv preprint arXiv:1607.06450}, 2016.

\bibitem{srivastava2014dropout}
Nitish {Srivastava}, Geoffrey {Hinton}, Alex {Krizhevsky}, Ilya {Sutskever},
  and Ruslan {Salakhutdinov},
\newblock ``Dropout: a simple way to prevent neural networks from
  overfitting,''
\newblock {\em Journal of Machine Learning Research}, vol. 15, no. 1, pp.
  1929--1958, 2014.

\bibitem{povey2011kaldi}
Daniel Povey, Arnab Ghoshal, Gilles Boulianne, Lukas Burget, Ondrej Glembek,
  Nagendra Goel, Mirko Hannemann, Petr Motlicek, Yanmin Qian, Petr Schwarz,
  et~al.,
\newblock ``The {Kaldi} speech recognition toolkit,''
\newblock in {\em IEEE ASRU}, 2011.

\bibitem{ghahremani2014pitch}
Pegah {Ghahremani}, Bagher {BabaAli}, Daniel {Povey}, Korbinian {Riedhammer},
  Jan {Trmal}, and Sanjeev {Khudanpur},
\newblock ``A pitch extraction algorithm tuned for automatic speech
  recognition,''
\newblock in {\em 2014 IEEE International Conference on Acoustics, Speech and
  Signal Processing (ICASSP)}, 2014, pp. 2494--2498.

\bibitem{sennrich2016BPE}
Rico {Sennrich}, Barry {Haddow}, and Alexandra {Birch},
\newblock ``Neural machine translation of rare words with subword units,''
\newblock in {\em ACL}, 2016, pp. 1715--1725.

\bibitem{watanabe2018espnet}
Shinji Watanabe, Takaaki Hori, Shigeki Karita, Tomoki Hayashi, Jiro Nishitoba,
  Yuya Unno, Nelson Enrique~Yalta Soplin, Jahn Heymann, Matthew Wiesner, Nanxin
  Chen, Adithya Renduchintala, and Tsubasa Ochiai,
\newblock ``{ESPnet}: End-to-end speech processing toolkit,''
\newblock {\em Proc. INTERSPEECH 2018}, p. 2207–2211, 2018.

\bibitem{paszke2017automatic}
Adam Paszke, Sam Gross, Soumith Chintala, Gregory Chanan, Edward Yang, Zachary
  DeVito, Zeming Lin, Alban Desmaison, Luca Antiga, and Adam Lerer,
\newblock ``Automatic differentiation in pytorch,''
\newblock in {\em NIPS 2017 Workshop Autodiff}, 2017.

\bibitem{kingma2015adam}
Diederik~P. {Kingma} and Jimmy~Lei {Ba},
\newblock ``Adam: A method for stochastic optimization,''
\newblock in {\em ICLR 2015 : International Conference on Learning
  Representations 2015}, 2015.

\bibitem{graves2006connectionist}
Alex {Graves}, Santiago {Fernández}, Faustino {Gomez}, and Jürgen
  {Schmidhuber},
\newblock ``Connectionist temporal classification: labelling unsegmented
  sequence data with recurrent neural networks,''
\newblock in {\em Proceedings of the 23rd international conference on Machine
  learning}, 2006, pp. 369--376.

\bibitem{he2016identity}
Kaiming {He}, Xiangyu {Zhang}, Shaoqing {Ren}, and Jian {Sun},
\newblock ``Identity mappings in deep residual networks,''
\newblock in {\em European Conference on Computer Vision}, 2016, pp. 630--645.

\end{thebibliography}

\end{document}